\documentclass[sigconf]{acmart}
\renewcommand\footnotetextcopyrightpermission[1]{}
\setcopyright{none}
\usepackage{blindtext}
\usepackage{enumitem}
\usepackage{xcolor}
\usepackage{threeparttable}
\usepackage{cleveref}
\usepackage{multirow}
\usepackage{booktabs}
\usepackage{makecell}

\begin{document}

\title{A Baseline Analysis for Podcast Abstractive Summarization}

\author{Chujie Zheng}
\email{chz@udel.edu}
\affiliation{%
  \institution{University of Delaware, USA}
}

\author{Harry Jiannan Wang}
\email{hjwang@udel.edu}
\affiliation{%
  \institution{University of Delaware, USA}
}

\author{Kunpeng Zhang}
\email{kpzhang@umd.edu}
\affiliation{%
  \institution{University of Maryland, USA}
}

\author{Ling Fan}
\email{lfan@tongji.edu.cn}
\affiliation{%
  \institution{Tongji University, China}
}

\renewcommand{\shortauthors}{Zheng et al.}

\begin{abstract}
  Podcast summary, an important factor affecting end-users' listening decisions, has often been considered a critical feature in podcast recommendation systems, as well as many downstream applications. Existing abstractive summarization approaches are mainly built on fine-tuned models on professionally edited texts such as CNN and DailyMail news. Different from news, podcasts are often longer, more colloquial and conversational, and noisier with contents on commercials and sponsorship, which makes automatic podcast summarization extremely challenging. This paper presents a baseline analysis of podcast summarization using the Spotify Podcast Dataset provided by TREC 2020. It aims to help researchers understand current state-of-the-art pre-trained models and hence build a foundation for creating better models.
\end{abstract}

\maketitle

\section{Introduction}
The podcast industry has been dramatically growing and gaining massive market appeal. For example, Spotify spent approximately \$200 million on the acquisition of Gimlet Media in 2019. However, the discovery and understanding of podcast content seem less progressive as compared to other types of media, such as music, movie, and news. This calls for more computationally effective methods for podcast analysis, including automatic summarization.

With the rapid development in Natural Language Processing, especially the success of attention mechanism and Transformer architecture \cite{vaswani2017attention}, the text summarization task has received increasing attention and many models have been proposed to achieve good performance, especially in the news summarization field \cite{lewis2019bart, raffel2019exploring, yan2020prophetnet}. They are all trained and tested using well-known CNN and DailyMail (CNN/DM) dataset where the headlines are served as the ground truth of summaries.

In this short paper, the dataset we study is the recently released TREC 2020 Spotify Podcasts Dataset \cite{clifton2020spotify}, which consists of 105,360 podcast episodes with audio files, transcripts (generated using Google ASR), episode summaries, and other show information.  Different from news, podcasts have unique characteristics, such as lengthy, multi-modal, more colloquial and conversational, and nosier with contents on commercials and sponsorship, which makes podcast summarization task more challenging. In this study, we aim to share our preliminary results on data preprocessing and some baseline analysis, which is expected to empirically show the aforementioned data specialty and build a foundation for subsequent podcast analyses. The code and pre-trained models will be released after the TREC 2020 competition \footnote{https://github.com/chz816/podcast-summarization-baseline}.

\section{Data Preprocessing}

The Spotify podcast dataset has 105,360 podcast episodes from 18,376 shows produced by 17,473 creators. The average duration of a single episode is 30 minutes, while the longest can be over 5 hours and the shortest is only 10 seconds. The TREC Podcast Track organizers form the "Brass Set" by cutting down the dataset to 66,245 podcast episodes using the following rules:

\begin{description}[font=$\bullet$~\normalfont\scshape\color{black!50!black}]
    \item Remove episodes with descriptions that are too long (> 750 characters) or too short (< 20 characters); 
    \item Remove "duplicate" episodes with similar descriptions (by conducting similarity analysis); 
    \item Remove episodes with descriptions that are similar to the corresponding show descriptions, which means the episode description may not reflect the episode content. 
\end{description}

On top of the Brass Set, we impose several extra constraints to form a cleaner dataset as follows:

\begin{description}[font=$\bullet$~\normalfont\scshape\color{black!50!black}]
    \item Remove episodes with emoji-dominated descriptions, i.e., descriptions with less than 20 characters after removing emojis.
    \item Remove episodes longer than 60 minutes to control the length of the episode descriptions. This constraint can be easily altered or relaxed if necessary. 
    \item Remove episodes with profanity language in the episode or show descriptions \cite{raffel2019exploring}.
    \item Remove episodes with non-English descriptions.
    \item Remove episodes with sponsorship/advertisement-dominated descriptions.
\end{description}

After preprocessing, the dataset has 24,250 episodes left, which serves the dataset for all analyses in this study (see Table \ref{data::preprocessing} for details).

\begin{table*}
\centering
\begin{tabular}{l|c}
\toprule
\multicolumn{1}{c|}{\textbf{Dataset Preprocessing}} & \multicolumn{1}{c}{\textbf{\# of Episodes}} \\
\hline
TREC Spotify Podcasts Dataset & 105360\\
After filtering by the TREC organizer (Brass Set)  & 66245 \\
After removing episodes with emoji-dominated descriptions & 56977 \\
After removing episodes longer than 60 minutes & 48074 \\
After removing episodes with profanity language & 33329 \\
After removing episodes with non-English descriptions & 32993 \\
After removing episodes with sponsorship/advertisement-dominated descriptions & 24250 \\
\toprule
\end{tabular}
\caption{Data Preprocessing and the Number of Episodes}
\label{data::preprocessing}
\end{table*}

\section{Baseline Models}
The abstractive summarization task aims to automatically generate the podcast episode summaries based on the episode transcripts. The ground truth is the summary written by the podcast creators. The performance of summarization models is often measured using the ROUGE score \cite{lin2004rouge}, particularly the F1 scores of ROUGE-1, ROUGE-2, and ROUGE-L \footnote{https://pypi.org/project/pyrouge/}. We also report recall (R) and precision (P).

We design two simple heuristic baselines for model comparisons:

\begin{description}[font=$\bullet$~\normalfont\scshape\color{black!50!black}]
\item [Baseline 1:] Select the first $k$ tokens from the transcript as the summary.
\item [Baseline 2:] Select the last $k$ tokens from the transcript as the summary.
\end{description}

The idea behind both baselines is that the beginning or the end of the podcast may contain more important content information. Their performance is shown in Table \ref{baseline model::performance}, with k being varied between 100 and 500. We choose the maximum value of k to 500 because BERT \cite{devlin2018bert} and other Transformer-based \cite{vaswani2017attention} models as we will discuss in the next section truncate the input to 512 tokens. The results exhibit an obvious pattern that longer summary tends to capture more words (measured by ROUGE-1) and phrases (measured by ROUGE-2 and ROUGE-L) that are also in the true summary, which often leads to higher recall but lower precision. The key takeaways are: (1). choosing k = 100 yields the best combined F1 score, which means 100 tokens (words) are long enough to capture the major summarization information. This is compatible with the distribution of the true summaries, where the average summary length is 44 and the maximal length is 144. (2). Baseline 1 has the highest F1 scores, which means the starting part of podcasts contains more useful and related information to podcast summaries than the ending part. This is also consistent with our observation that podcast episodes often give some overview at the beginning to tell the listeners what to expect.

\begin{table*}
\centering
\begin{tabular}{c|c|ccc|ccc|ccc}
\toprule
\multicolumn{2}{c|}{\multirow{2}*{\textbf{Model}}} & \multicolumn{3}{c|}{\textbf{ROUGE-1}} & \multicolumn{3}{c|}{\textbf{ROUGE-2}} & \multicolumn{3}{c}{\textbf{ROUGE-L}} \\
\cline{3-11}
\multicolumn{2}{c|}{~} & R & P & F & R & P & F & R & P & F\\
\hline
\multirow{5}*{Baseline 1} & k=100 & 38.07 &	\textbf{15.52} &	\textbf{21.11} &	8.18 &	\textbf{3.30} &	\textbf{4.49} & 33.20 &	\textbf{13.57} &	\textbf{18.44}  \\
~ & k=200 & 51.31 &	10.80 &	17.25 &	12.79 &	2.73 & 4.32 &	46.11 &	9.71 &	15.51   \\
~ & k=300 & 58.01 &	8.28 &	14.09 &	15.59 &	2.28 &	3.83 &	53.01 &	7.57 &	12.87   \\
~ & k=400 & 62.20 &	6.77 &	11.88& 	17.59 &	1.97 &	3.42 &	57.41 &	6.25 &	10.97   \\
~ & k=500 & \textbf{65.17} &	5.76 &	10.31 &	\textbf{19.21} &	1.76 &	3.10 &	\textbf{60.60} &	5.36 &	9.59  \\
\hline
\multirow{5}*{Baseline 2} & k=100 & 31.88 &	 \textbf{13.08} &	 \textbf{17.76} &	 3.88 &	\textbf{1.62} &	2.18 &	27.63 &	\textbf{11.39} &	\textbf{15.43}  \\
~ & k=200 & 44.25 &	9.32 &	14.89 &	6.58 & 1.42 &	\textbf{2.23} &	39.43 &	8.32 &	13.27   \\
~ & k=300 & 51.04 &	7.29 &	12.39 &	8.65 &	1.29 &	2.14 &	46.22 &	6.61 &	11.22   \\
~ & k=400 & 55.45 &	6.04 &	10.59 &	10.42 &	1.20 &	2.04 &	50.78 &	5.53 &	9.70   \\
~ & k=500 & \textbf{58.68} &	5.19 &	9.29 & \textbf{11.93} &	1.13 &	1.95 &	\textbf{54.18} &	4.79 &	8.57   \\
\toprule
\end{tabular}
\caption{Model Performance for Two Baseline Models}
\label{baseline model::performance}
\end{table*}

\section{SOTA Model Experiments}
In this section, we conduct a number of experiments for the podcast summarization task using three current state-of-the-art (SOTA) summarization models, including BART \cite{lewis2019bart} \footnote{In this paper, we use DistilBART provided by Hugging Face. It achieves better performance than the original BART model in our experiment.}, T5 \cite{raffel2019exploring}, and ProphetNet \cite{yan2020prophetnet}. More specifically, we use the pre-trained models, fine-tune them using the news datasets (CNN and DailyMail datasets \cite{nallapati2016abstractive}), and the preprocessed podcast dataset from Section 2. The goal is to get an overview idea about the performance of the SOTA models, which builds a foundation for better model innovation.  All experiments are conducted under a machine with two Tesla V100 GPUs.

We split our processed podcast dataset into training, validation and testing sets by 60\%, 20\%, and 20\% at random, resulting in 14,550 observations in the training set and 4850 observations in both validation and testing sets. Based on the baseline analysis in the previous section, we choose the beginning part of the episode transcripts as the input (we use the default settings that use 1024 tokens for BART and T5 and 512 tokens for ProphetNet) and the episode description from creators as the summarization ground truth. Table \ref{experiment::compare} shows the experiment results, from which we have the following observations. 

(1). The performance of the SOTA models is comparable to the baseline models, which indicates that there is plenty of headroom for improvements and calls for more research in this emerging area.

(2). The F1 scores for ROUGE 1, 2, and L of ProphetNet on the CNN/DM dataset are 44.20, 21.17, 41.30, but the corresponding best F1 scores in Table 3 for the podcast dataset are only 26.76, 7.95, and 22.71. This huge performance gap implies that the podcast summarization task could be more challenging than the news headline summarization task due to the podcast's unique characteristics aforementioned. 

(3). Fine-tuning the pre-trained models on the CNN/DM dataset for podcast summarization may result in lower performance compared with the vanilla pre-trained models, e.g. BART and ProphetNet. This urges us to think more about the lexicon differences between the podcast dataset and other existing datasets used in summarization tasks, such as CNN/DM, Gigaword \cite{rush2015neural}, BigPatent \cite{sharma2019bigpatent}, and PubMed\cite{cohan2018discourse}.

We also provide some sample generated podcast summaries from different models in our repository. 

Based on the baseline analysis in this paper, we discuss a number of directions for future research:

\begin{description}[font=$\bullet$~\normalfont\scshape\color{black!50!black}]
    \item Summarization based on long narrative structure: as discussed in \cite{papalampidi2020screenplay}, simple position heuristics are not sufficient for long narratives (such as podcast transcripts) summarization. How to define a narrative structure for better podcast summarization is interesting and worthy of the topic.
    \item Conversation summarization: podcasts are often conversational, colloquial, and multi-people. How to leverage existing research such as \cite{ganesh2019abstractive, zhu2006summarization, sood2013topic} to help podcast summarization is still largely missing.
    \item Multi-modal podcast analysis: the audio files of podcasts contain much richer information than the text transcripts, such as music, emotion, pitch, etc. We believe the multi-modal analysis is critical for podcast understanding and thus should play an important role in podcast summarization and recommendation \cite{baltruvsaitis2018multimodal}. 
    \item Long-document transformer: how to leverage recent research on \cite{beltagy2020longformer}, and \cite{kitaev2020reformer} to potentially use the full podcast transcripts during training. 
\end{description}

\begin{table*}
\centering
\begin{threeparttable}
\begin{tabular}{l|ccc|ccc|ccc}
\toprule
\multicolumn{1}{c|}{\multirow{2}*{\textbf{Model}}} & \multicolumn{3}{c|}{\textbf{ROUGE-1}} & \multicolumn{3}{c|}{\textbf{ROUGE-2}} & \multicolumn{3}{c}{\textbf{ROUGE-L}} \\
\cline{2-10}
\multicolumn{1}{c|}{~} & R & P & F & R & P & F & R & P & F\\
\hline
Baseline 1 (k=100) & \textbf{38.07} &	\textbf{15.52} &	\textbf{21.11} &	\textbf{8.18} &	\textbf{3.30} &	\textbf{4.49} & \textbf{33.20} &	\textbf{13.57} &	\textbf{18.44} \\
Baseline 2 (k=100) & 31.88 &	13.08 &	 17.76 &	 3.88 &	1.62 &	2.18 &	27.63 &	11.39 &	15.43 \\
\hline
DistilBART \cite{lewis2019bart} \tnote{1} & 30.02 &	19.44 & 22.26 &	6.26 &	4.20 &	4.73 & 26.05 & 16.98 & 19.39\\
DistilBART \cite{lewis2019bart} + CNN/DM \tnote{*} & 26.50 &	20.76 &	22.05 &	5.15 &	4.05 &	4.27 &	23.02 &	18.14 &	19.21 \\
DistilBART \cite{lewis2019bart} + Podcast \tnote{**} & 32.36 & \textbf{25.44} &	\textbf{26.76} &	9.28 &	\textbf{7.31} &	7.67 &	27.36 &	\textbf{21.67} &	\textbf{22.71} \\
T5 \cite{raffel2019exploring} \tnote{2} & 25.74 &	19.39 &	20.59 &	4.75 &	3.52 &	3.75 &	22.14 &	16.80 &	17.77 \\
T5 \cite{raffel2019exploring} + CNN/DM \tnote{*} & 31.26 &	17.09 &	21.03 &	5.90 &	3.19 &	3.93 &	26.95 &	14.82 &	18.18 \\
T5 \cite{raffel2019exploring} + Podcast \tnote{**} & 31.66 &	18.43 &	22.15 &	6.46 &	3.72 &	4.46 &	24.91 &	14.59 &	17.49 \\
ProphetNet \cite{yan2020prophetnet} \tnote{3} & 20.78 & 22.08 &	19.52 &	6.23 &	6.75 &	5.84 &	17.90 &	18.65 &	16.59\\
ProphetNet \cite{yan2020prophetnet} + CNN/DM \tnote{*} & 32.52 &	13.60 &	17.85 &	9.13 &	3.59 &	4.77 &	28.29 &	11.51 &	15.20\\
ProphetNet \cite{yan2020prophetnet} + Podcast \tnote{**} & \textbf{34.26} &	19.01 &	22.61 &	\textbf{12.37} &	6.66 &	\textbf{7.95} &	\textbf{29.57} &	15.95 &	19.12\\
\toprule
\end{tabular}
\begin{tablenotes}
	\item[1] DistilBART: we use Hugging Face Transformers (model: sshleifer/distilbart-cnn-12-6)
	\item[2] T5: we use Hugging Face Transformers (model: t5-small)
	\item[3] ProphetNet: we use released ProphetNet-large-160GB checkpoint
	\item[*] Fine-tuned on CNN/DM Dataset
	\item[**] Fine-tuned on Podcast Dataset
\end{tablenotes}
\end{threeparttable}
\caption{Performance Comparison of Different Models}
\label{experiment::compare}
\end{table*}

\section{Conclusion}
In this paper, we present the performance of podcast summarization using two baselines and SOTA models on the Spotify podcast dataset. We discuss several directions for future research in this field. We hope this pioneering baseline analysis and implementation can help researchers make more much-needed innovation in this exciting emerging research area.


\bibliographystyle{ACM-Reference-Format}
\bibliography{bio}

\end{document}